\RequirePackage{fix-cm}
\documentclass[twocolumn]{svjour3}          
\smartqed  
\usepackage{graphicx}
\usepackage{hyperref}
\usepackage{cite}

\begin{document}
\title{Image Conditioned Keyframe-Based Video Summarization Using Object Detection}

\author{Neeraj Baghel         \and
        Suresh C. Raikwar   \and
        Charul Bhatnagar
}
\institute{Neeraj Baghel
           \and
           Suresh C. Raikwar 
              \and
            Charul Bhatnagar 
              \at
              Department of Computer Engineering and Applications, GLA University, Mathura, India \\
            \email{nbaghel777@gmail.com}
}
\date{Received: date / Accepted: date}
\maketitle
\begin{abstract}
Video summarization plays an important role in selecting keyframe for understanding a video. Traditionally, it aims to find the most representative and diverse contents (or frames) in a video for short summaries. Recently, query-conditioned video summarization has been introduced, which considers user queries to learn more user-oriented summaries and its preference. However, there are obstacles in text queries for user subjectivity and finding similarity between the user query and input frames. In this work, (i) Image is introduced as a query for user preference (ii) a mathematical model is proposed to minimize redundancy based on the loss function \& summary variance and (iii) the similarity score between the query image and input video to obtain the summarized video. Furthermore, the Object-based Query Image (OQI) dataset has been introduced, which contains the query images. The proposed method has been validated using UT Egocentric (UTE) dataset. The proposed model successfully resolved the issues of (i) user preference, (ii) recognize important frames and selecting that keyframe in daily life videos, with different illumination conditions. The proposed method achieved 57.06\% average F1-Score for UTE dataset and outperforms the existing state-of-the-art by 11.01\%. The process time is 7.81 times faster than actual time of video Experiments on a recently proposed UTE dataset show the efficiency of the proposed method.

\end{abstract}

\newpage
\section{Introduction}
The video-capable mobile devices becoming increasingly ubiquitous. There is an analogous increase in the amount of video data that is captured and stored. Additionally, as the difficulty of capturing video, the cost of storage decreases and tends to see a corresponding increase in the quality of captured videos. As a result of this, it becomes very difficult to watch or discover interesting video clips among the vast amount of data. One solution to this problem is to lies in the development of a video summarization system, which can automatically locate these interesting clips and generate a final curated video summary.

In current years, interest in video summarization is increased due to a large amount of video data. Certainly, professionals and consumers both have access to video retrieval nowadays. The image contains a large amount of data whereas the video is a collection of images, which contains huge information and knowledge. Thus, it is very hard for users to watch or discover the incidents in it. Quick video summarization methods enable us to speedily scan a long video by removing irrelevant and redundant frames.

The video summarization is a tool for creating a compact summary of a video. It can either be an order of stable images (keyframes) or motion pictures (video skims)  \cite{02kansagara2014study}. The video can be summarized using (i) keyframes, and (ii) video skims.

The keyframes are used to represent important information contained in video. These are also named as R-frames, representative frames, still-image synopses and a collection of prominent images obtained from the video data. The challenges in selecting keyframes are as follows. (i) Redundant frames are selected as a keyframe. (ii) Difficult to make a cluster when content is non-identical.
The video skims represents moving story borad of a video \cite{29truong2007video}. The video is split into many portions that are video clips with a smaller length. Every portion follows a part of a video or a regular result. The trailer of a movie is an example of video skimming.

Finally, different viewers will have different preferences on what they find interesting in a video. With traditional hand-editing, viewers only see the segments deemed interesting by the editor, and the time cost for a human editor to create multiple edits of a single video is significant. The ability to generate multiple possible summaries rather than just a single summary would be a very useful feature for a video summarization system to have, as would be the ability to learn a specific user’s preferences over time.

The proposed framework has taken an image query for user preference and uses both global \& local features to learn user preference from that image query to generate video summary efficiently. Object detection is used major user preference based on objects as local features while a salient region is used for focused area \& colours in an image as global features.

\begin{figure}
\includegraphics[scale=.57]{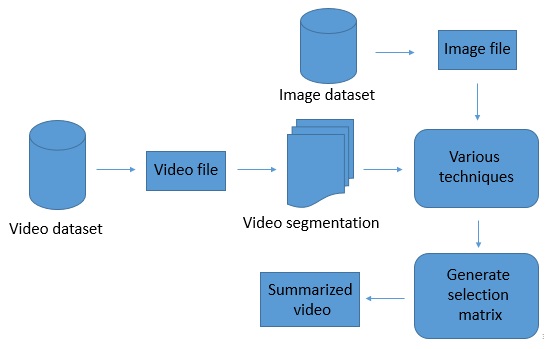}
\caption{ Video summarization workflow: Video and image are selected from their respective datasets. Further various techniques are applied to generate the selection matrix. On behalf of the selection matrix summarised video is formed.}
\end{figure}

The objective of the proposed work is to resolve the issue of the wrong selection of keyframe with a mathematical model to minimize redundancy based on similarity score between query image and frames of the input video to obtain the summarized video. Hence, the proposed method has used the following features to measure the degree of similarity between an input query image and video. (i) Local features based on object detection and its details. (ii) Global features based on salient regions. 

The primary contributions of the presented work are:\begin{enumerate}
    \item A mathematical method to calculate summary variance to reduce redundancy between frames.
    \item Defined the loss function for obtaining a selection score for frames in a video.
    \item The Object-based Query Image (OQI) dataset is prepared for the selection of a query image.
    \item An adaptive method to compute threshold is defined for the selection of keyframes using standard deviation. 
     \end{enumerate} 

In addition to this primary contribution, parts of the proposed work also serve as relevant contributions in isolation. These include: Perform video summarization on commodity hardware.

The remaining structure of this paperwork is as follows: Section-2 provides the related work of video summarization. Section-3 describes the Datasets used and its setting. Section-4 describes the proposed mathematical model of the problem and its solution by calculating loss function. Also, Flowchart and its working consist of video segmentation, feature extraction, frame scoring, and summary generation. Section-5 describes the Implementation Details, evaluation metrics, experimental results obtained by the proposed model. Section-6 discuss about GUI tool for summary generation. In the last, Section-7 discussion about the conclusion of the proposed model. 

\section{Related Work}
Although video summarization has been an active research topic in computer vision nowadays. The high computational capabilities of modern hardware allow us to process a video in a fraction of the time if required, which when combined with the evolution of modern vision techniques such as deep neural networks, has resulted in a significant increase in the breadth of techniques which are viable to apply to the topic of video summarization. Combined with the vast quantity of prior work involving video summarization, many interesting research prospects are available to pursue.

Although the primary focus is video summarization, some of the steps performed during the proposed work need a significant amount of prior research. Among these query-focused video summarizing, which simply deals with taking a video and dividing it up into a number of segments, and image and video feature extraction related to the query, which deals with extracting relevant and useful features from images and videos.

In, \cite{24sharghi2017query} author has proposed a new technique on video summarization, which is query-focused  \cite{37zhang2016summary}, that includes user perspective based on text queries related to video into the summarization method. This is a promising way to personalize video summary. The author collects user annotations and meets the issues of good evaluation measures for system-generated summaries to user-labeled summaries. Author’s contribution, (i) collect dense tags for the dataset, (ii) a memory network using a sequential determinantal point process \cite{14lin2015summarizing} query-focused video summarization. The author uses the memory network to take user queries for a video within each shot onto different frames.

In \cite{27zhang2018query} query-conditioned summarization is introduced using a three-player generative adversarial network. where generator is used to learn the joint representation of query and video. The discriminator takes three different summaries as input and discriminate the real summary from the other two summaries which are randomly generated.

Generic video summarization \cite{11khosla2013large,12kim2014joint,14lin2015summarizing,37zhang2016summary,38zhang2016video}, has been studied for global keyframes and efficient analysis of video. For shot-level summarization  \cite{14lin2015summarizing,15lu2013story,25song2015tvsum,39yao2016highlight}, Song et al. \cite{25song2015tvsum} finding important shots using learning of visual concepts shared between videos and images. In  \cite{39yao2016highlight}, distinguish highlight segments from non-highlight ones by using a pair-wise deep ranking model. In frame level video summarization \cite{04gong2014diverse,11khosla2013large,12kim2014joint,40zhang2018query}, Khosla et al. \cite{11khosla2013large} use web based images before video summarization. In  \cite{04gong2014diverse},probabilistic model is used for learning of sequential structures and generating further summaries. Object-level video summarization \cite{17meng2016keyframes,26zhang2018unsupervised}  extracts objects to perform summarization. Also, there are two GAN-based networks \cite{16mahasseni2017unsupervised,40zhang2018query} that include adversarial training. However, user preferences is not considered, so summaries may not generalize and robust for different users. Therefore, the video summarization based on query-conditioned came into focus and provide more personalized summary to user. 

\begin{table}
\caption{Related work to video summarization}
\includegraphics[scale=0.65]{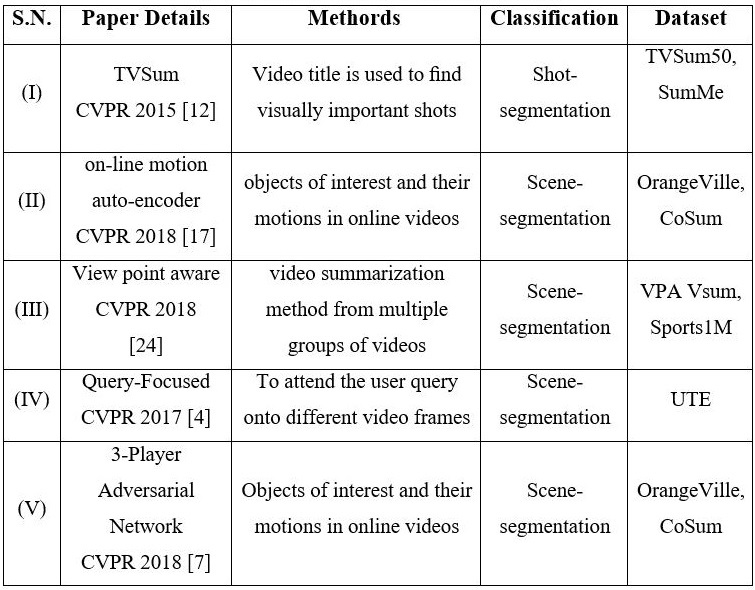}
\end{table}

Query-conditioned video summarization  \cite{10ji2017query,19oosterhuis2016semantic,23sharghi2016query,24sharghi2017query,41vasudevan2017query} takes queries given by user into consideration which are in the form of texts to learn and generate user-oriented summaries. To tackle this challenge \cite{23sharghi2016query} trained a Sequential and Hierarchical DPP (SH-DPP). In \cite{41vasudevan2017query}, to pick relevant and representative frames adopt a quality-aware relevance model.  

Specifically, Oosterhuis et al. \cite{19oosterhuis2016semantic} generate visual trailers based on graph-based method for selecting most relevant frames to a given user query. Ji et al.  \cite{10ji2017query} formulate incorporating web images task obtained from user query searches. 

Recently, Sharghi et al. \cite{24sharghi2017query} Instead of using generic task datasets for query conditioned task, they propose a new dataset, its evaluation metric and a technique based on this new dataset. They propose an adversarial network that summarize videos based on user queries and does not rely on external source (web images).

\section {Datasets and Settings}

In the Proposed methodology, experiments are accomplished on the prevailing UT Egocentric (UTE) dataset. The dataset contains four videos each 3.5 hours long, completely different uncontrolled daily life situations. A lexicon is provided for user queries consists of various set of forty-eight ideas, based on daily life. As for the queries, four completely different situations videos are enclosed to formalize comprehensive queries. Also follow three scenarios which are (i) queries wherever all ideas seem along within the same video shot, (ii) queries where all ideas seem however not within the same shot, and (iii) queries wherever just one of the concepts seem.

\begin{table}
\caption{Dataset used for experiments}
\includegraphics[scale=0.6]{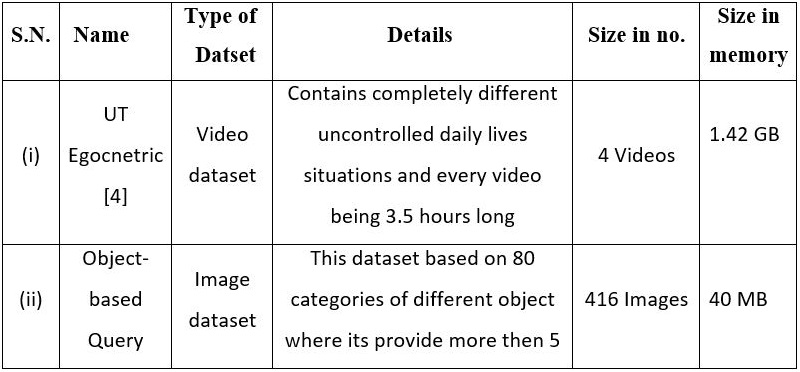}
\end{table}

Also, introduced conjointly created a tiny Object-based Query Image (OQI) dataset of pictures for the testing image as a question and object detection and feature extraction. During this dataset, various pictures of totally different eighty categories. Every category has quite five pictures \& images will have quite one totally different objects. The proposed dataset got impressed by the COCO dataset  \cite{03lin2014microsoft} and conjointly contained pictures are taken by us.

\section{Proposed method}
Video summarization using object detection is a great challenge in the field of artificial intelligence for the machines as well as the programmers to train machines in such a way that it recognizes the keyframes automatically. The proposed method facilitates query-conditioned video summarization by extracting an Object's detail (local feature) and Salient region (global feature) feature extraction. It takes the Image as a query into consideration with its features recognizes that same feature into the video frame. The framework of the proposed approach is shown in Fig.3. The first part extracts objects and its visual feature, in order to provide comprehensive summary for the given image.

Additionally, introduced a small Object-based Query Image dataset of Images for the testing image-query. In this dataset, there are images of 80 different classes. Each class has more than 5 images \& image can have more than one different objects. This dataset has inspired by Common Objects in Context (COCO) dataset  \cite{03lin2014microsoft} and also contained some other images which are taken by us. The proposed method has also tested these on different videos and images that show this proposed model not restricted query image and videos to these datasets.

\subsection{  Mathematical Model }

\begin{figure}
\includegraphics[scale=.7]{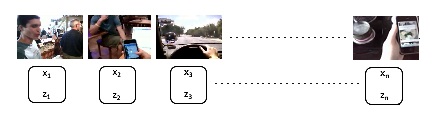}
\caption{Mathematical model: where x represents a feature of that frame and z represents the selection value of that frame.}
\end{figure}

Consider $X = [x_{1}, x_{2}, x_{3}, ..., x_{n}]$ be a feature matrix for a video with n segment (frame) features x. The feature representation \cite{32kanehira2018viewpoint} of the summary for video defined as 
\begin{equation} V_s = X^{T} Z \end{equation} 
Where Z is selection matrix $ Z=[z_1, z_2, z_3, ..., z_n] $ and $ z_{n}  \in \{0, 1\}^T$  is the selection variable. The model \cite{32kanehira2018viewpoint} is based on view point in multiple videos. Loss function is defined by diverse content, same group videos and different group videos whereas proposed method used different distance functions based on similarity between image query and input video.

\textbf{Objective}: To find the values of $z_{n}$ in order to minimize the loss function L(Z)
\begin{equation}
L(Z)=\lambda_1 Tr(S^v) - \lambda_2 Tr(S^q) 
\end{equation}
Where $S^v$ is Summary variance, $S^q$ is Distance score, Tr is Trace function and $\lambda_1 \&  \lambda_2 $ are parameters to control the importance of each term. So, variance $S^V$ for the summary of a video is defined as 
\begin{equation}
S^V = \sum_{i=1}^n z_i(x_i-V_s)(xi-V_s)^T
\end{equation} 
Thus, by using (1), trace of (3) can be written as: 
\begin{equation}
  Tr(S^v) = \sum_{i=1}^n z_i x_i^T x_i - z_i^T x_i x_i^T z_i  
\end{equation}
By placing all n frames together using stacked variable it can be written as: 
\begin{equation}
  Tr(S^v) = Z^T (P-Q) Z  
\end{equation}
Where $Z=[z_1, z_2, z_3, ...,z_n]$, $P= diag(X^T,X)$ and $Q=XX^T$ Distance Matrix D is defined as 
\begin{equation}
 D= [d_1, d_2, d_3, ..., d_n]   
\end{equation}
Where d is the Cumulative Distance Function defined as
\begin{equation}
d_i=	\phi_1(f^q_1 , f^v_1) + \phi_2(f^q_2 , f^v_2) + ...+ \phi_m(f^q_m , f^v_m)   
\end{equation}
Where $f^q$ is a feature of the query image, $f^v$ is a feature of video frames and  $\phi$  is Distance Function, m is feature no. Therefore, Distance score $S^q$ for the summary of a video can be defined as
\begin{equation}
  S^q =DD^T Z   
\end{equation}
Thus, its trace can be written as 
\begin{equation}
Tr(S^q) = R Z    
\end{equation}
Where, R = diag(D,D\^T)

By putting the trace values of Summary Variance $Tr(S^v)$ (5) and Distance Score $Tr(S^q)$ (9) in the Loss function L(Z). 

\begin{equation}
    L(Z)= Z^T (\lambda_1P) Z - Z^T (\lambda_1Q+\lambda_2R) Z
\end{equation}
For solving eq(10) used method same \cite{32kanehira2018viewpoint} Given the loss function represented by L(Z) = f(Z)-g(Z) where $f(\cdot)$ and $g(\cdot)$ are convex functions, By utilized a well-known CCCP (concave-convex procedure) algorithm \cite{33yuille2002concave,34yuille2003concave} to solve it. Where f(Z) is $ Z^T (\lambda_1P) Z$ and g(Z) is $Z^T (\lambda_1Q+\lambda_2R) Z$. In this, the loss function can be decomposed into the difference of two convex functions. In $t^{th}$ iteration, it will converge the values of Z which lies between 0 and 1 as denoted as $Z_m$.

In $Z_m$ Values lies between 0 and 1 where values close to 1 are for important frames, while close to 0 represent unrelated frame. Then threshold is required for selecting important frames calculated as the standard deviation as the values are Gaussian distributed. Convert $Z_m$ to Z on the basis of adaptive threshold value.  Values having greater then threshold values will be considered as 1 while rest are 0. Therefore selection matrix Z is defined as  
 \begin{equation}
     Z   = Z_m > \sigma 
 \end{equation}   
 Where $\mu$ is a standard deviation from $Z_m=G(\mu, \sigma)$. At last, generated summarized video by accumulated keyframe defined by selection 
matrix which can be defined as: $V_s = X^T Z  (Eq 1)$ where, X is feature matrix and Z is selection matrix. 

\subsection{Further the flowchart of the proposed model }
The flowchart of the proposed model of video summarization is shown in Fig.3. The proposition is broadly classified into these phases: (a) video segmentation, (b) feature extraction, (c) frame scoring, and (d) summary generation. 

\begin{figure}
\includegraphics[scale=.5]{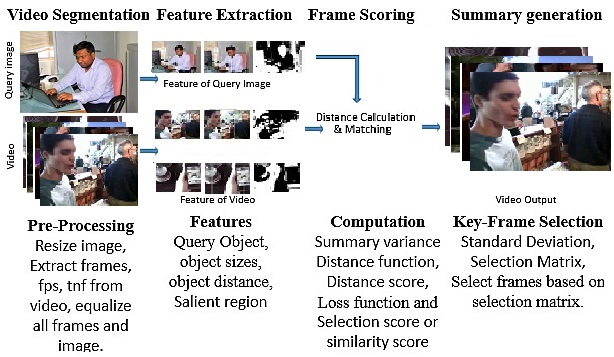}
\caption{Framework for video summarization: contains four major steps video segmentation, feature extraction, frame scoring and summary generation. }
\label{Famework}
\end{figure}

\subsubsection{Video Segmentation }
Video Segmentation is the significant step in almost all video processing systems, where for a target input video V, a segmentation $v_n$ is generated. This phase includes a pre-processing step to eliminate undesirable frames. In the pre-processing step, (i) frames are extracted from the input video, (ii) histogram equalization is applied, (iii) resizing of images to a default size, (iv) removing an unnecessary part from all images, (v) extract frames-per-second and (vi) total-no-of-frames from the input video. Each video of UTE dataset V is partitioned into ‘n’ frames (5-seconds long) for fair comparison with related work \cite{24sharghi2017query}, where ‘n’ is the total number of frames. Casings (5-second long) is utilized outline extraction technique to compare with existing work. 
\begin{figure}
\includegraphics[scale=.38]{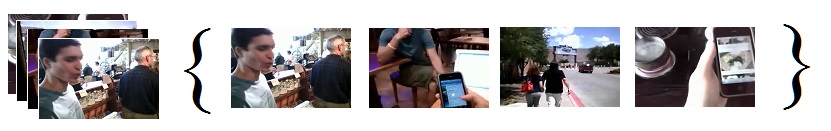}
\caption{Video segmentation: frames are extracted from selected video and various pre-processing techniques are applied.}
\end{figure}

\subsubsection{Feature extraction }
Feature extraction methods should be video summary oriented. The  features are first extracted from the query image and each frame of input video, then aggregated within each segment $x_n$ to obtain a feature vector X. The proposed framework has taken both global and local features to generate video summary effectively. For local features, the proposed work utilized object detection. The objective of object detection is to recognize all occurrences of objects from a recognized classification, similar to people, vehicles or faces in an image. An item extraction module is intended to provide the thing division strategy that furthermore gives a dependable contribution to the pursuit instrument. The proposed method has utilized You Only Look Once (YOLO) method \cite{28redmon2018yolov3} with 80 different classes for recognition, and trained on COCO dataset \cite{03lin2014microsoft}. 

The classes considered in the proposed method are: [‘person’, ‘umbrella’, ‘tie’, ‘backpack’, ‘handbag’,  ‘suitcase’, ‘bicycle’, ‘motorcycle’, ‘bus’, ‘truck’,  ‘car’, ‘airplane’, ‘train’, ‘boat’, ‘traffic\_light’,  ‘stop\_sign’, ‘bench’, ‘fire\_hydrant’, ‘parking\_meter’, ‘bird’, ‘dog’, ‘sheep’, ‘elephant’, ‘zebra’, ‘cat’, ‘horse’, ‘cow’, ‘bear’, ‘giraffe’, ‘frisbee’,  ‘snowboard’, ‘kite’, ‘baseball\_glove’, ‘surfboard’,  ‘skis’, ‘sports\_ball’, ‘baseball\_bat’, ‘skateboard’, ‘tannis\_racket’, ‘bottle’, ‘cup’, ‘knife’, ‘bowl’,  ‘wine\_glass’, ‘fork’, ‘spoon’, ‘banana’, ‘sandwich’,  ‘broccoli’, ‘hot\_dog’, ‘donut’, ‘apple’, ‘orange’,  ‘carrot’, ‘pizza’, ‘cake’, ‘chair’, ‘potted\_plant’, ‘dining\_table’, ‘couch’, ‘bed’, ‘tv’, ‘toilet’,  ‘mouse’, ‘keyboard’, ‘laptop’, ‘remote’, ‘cell\_phone’, ‘toaster’, ‘microwave’, ‘refrigerator’, ‘oven’, ‘sink’,  ‘book’, ‘vase’,   ‘teddy\_bear’, ‘toothbrush’, ‘clock’, ‘scissors’,  ‘hair\_drier’] 

The YOLO \cite{28redmon2018yolov3}, is one of the faster object detection methods. The proposed method has used the trained model on the COCO dataset \cite{03lin2014microsoft}. The output for every frame Using YOLO consists of three lists (each representation are 84-dimensional vectors), YOLOv3 makes predictions at three scales, by down-sampling the size of the input image into blocks of size 32X32, 16X16 and 8X8 blocks. However, the proposed method has used only 16X16 and 8X8 blocks because very small objects are not require.  

\begin{figure}
\includegraphics[scale=.6]{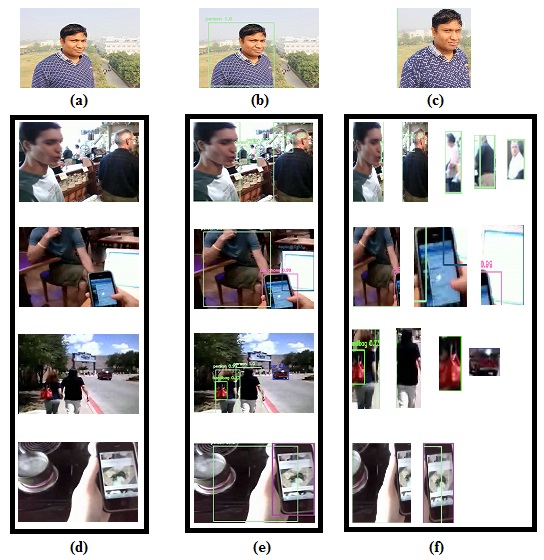}
\caption{Local feature extraction (a) processed query image (b) object detection on query image (c) object extracted from query image (d) processed video frames (e) object detection on video frames (f) object extracted from video frame.}
\end{figure}

If the prediction confidence score is generated if more than 0.5 then the object is extracted from that frame. Then the output of these two feature vector and object category is matched with other frames of a given video. With confidence threshold 0.5 and non-maximal suppression threshold 0.4. Finally, the proposed method uses these various object properties as local features. For global features, the proposed method is using the salient region. The objective of the salient region is to find important regions in the video frames, color combination and depth of that region. Recognize the important region of image from the HSV color model. Author \cite{2012regionsinterest} has used HSV color model to find salient region in stereo images by using subtraction of left and right images with saturation and value components of HSV color space. To obtain a salient region in mono image proposed methord used the HSV color model using functions. 
\begin{equation}
     Sr = Exp(-(V - S)) > \alpha
\end{equation}  
Where V is value plane, S is the saturation plane, $\alpha$ is parameter threshold to select the region, Exp is exponential function and Sr is the salient region. The value of $\alpha$  is always lie between 0 to 1 and the value is as much close to 1 function select more salient pixel from the image. However, it may skip some pixel which may salient but due to some noise, its value is less. The value of $\alpha$ empirically by testing this on various images of GQI dataset  to be ‘0.7’.   

\begin{figure}
\includegraphics[scale=.56]{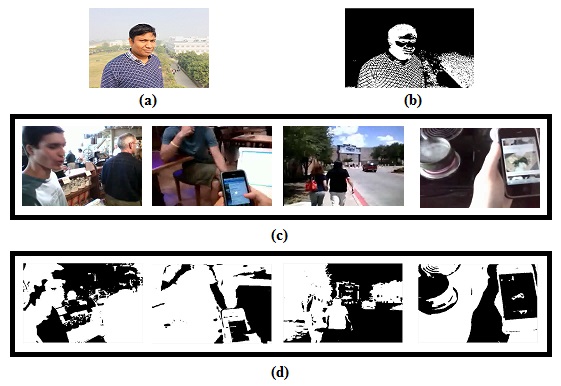}
\caption{Global feature extraction (a) processed query image (b) salient region on query image (c) processed video frame (d) salient region on video frame.}
\end{figure}

\subsubsection{Frame Scoring }
In frame scoring, various steps are used to generate the similarity score of the input video frame. First, the step is calculating cumulative distance between the query image and input video. A cumulative distance is a collection of various distance calculated between features of the query image and frames of the input video. The proposed work have four distance function for (i) different objects (ii) location of the objects (iii) size of the objects (iv) salient region.

Denoted $d_i$ as the cumulative distance of $i^{th}$ frame, Distance function as $\phi$, a feature of query image as $f^q$ and  feature of the input video frame as $f^v$. Also proposed the cumulative distance function eq7 as $di= \phi_1(f^q_1-f^v_1) , \phi_2(f^q_2-f^v_2) , \phi_3(f^q_3-f^v_3) , \phi_4(f^q_4-f^v_4)$ where $\phi_1(f^q_1-f^v_1)$ is an absolute value of the difference between the number of different objects in the query image and video frame, $\phi_2(f^q_2-f^v_2)$ is a summation of the difference between location of a similar object in the query image and video frame, $\phi_3(f^q_3-f^v_3)$ is a summation of the difference between size of a similar object in the query image and video frame and then divide by total number of pixel and $\phi_4(f^q_4-f^v_4)$ is a summation of difference between a salient region of the query image and video frame and then divide by total no of pixel. 

The second step is to generate a distance matrix with the collection of cumulative distance function with respect to every frame of the input video. Distance matrix (D) is described as: $ D= [d_1, d_2 , d_3, ..., d_n]$ Where, d is the cumulative distance between the query image and input video.

The third step is to calculate the values of P, Q and R as described: $P = diag(X^T X) $, $Q = XX^T$ and $R = DD^T$ where diag is a diagonal matrix, X is a feature matrix, D is a distance matrix.

The fourth step is to find values of selection score $Z_m$ which lies between 0 and 1 with the help of Convergence of Concave-Convex Procedure  by minimizing the optimized equation of  loss function L(S)  in $t^{th}$ iteration. 

\begin{figure}
\includegraphics[scale=.6]{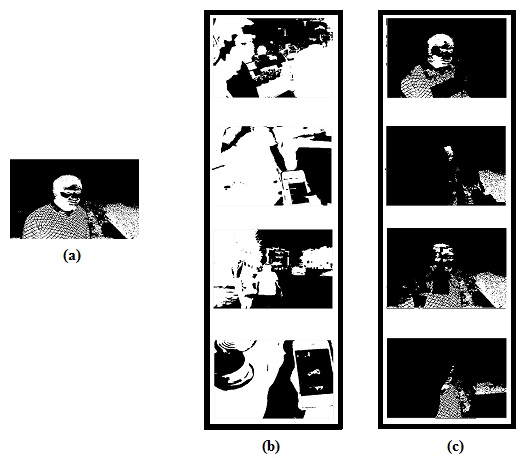}
\caption{Salient region difference: (a) salient region on query image (b) salient region on video frame (c) difference of salient region between query image and video frames.}
\end{figure}

\subsubsection{Key-Frame Selection }
Keyframe selection is the last step in video summarization where the important frame is selected from input video and generates the output video of those frames. A keyframe is selected on the basis of similarity score. In the proposed method, first selection matrix is generated, which takes the decision of selection of keyframe. The selection matrix is generated on the basis of the selection score where a threshold is applied to the selection score. If  the value is greater than the threshold then considered as keyframe otherwise that frame is discarded. The threshold is calculated as the standard deviation as the values are Gaussian distributed. Therefore, selection matrix z is defined as $Z   = Z_m > \sigma$  where $\sigma$ is a standard deviation from $Z_m=G(\mu, \sigma)$. 

At last, summarized video is generated by accumulated keyframe defined by selection matrix which can be defined as  $V_s = X^T Z$  where X is a feature matrix and Z is a selection matrix. 

\begin{figure}
\includegraphics[scale=.6]{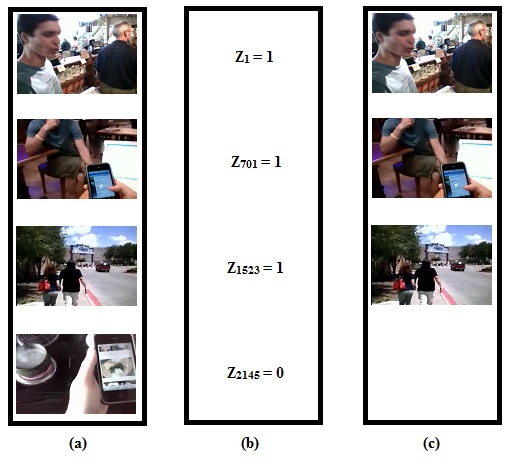}
\caption{Keyframe selection: (a) video frames (b) selection matrix (c) selected keyframe}
\end{figure}

\section{Experimental Results and Discussion}
In this chapter, the proposed method has been validated the approach using UT Egocentric (UTE) dataset. The proposed method successfully resolved the issues of (i) user preference, (ii) recognize important objects in frames according to user preference  and selecting keyframe in daily life videos, with different illumination conditions. Experiments is performed on UT Egocentric dataset shows the efficiency of the proposed method.

\subsection{Implementation Details }
The proposed video summarization system is implemented in PYTHON 3.6 at 1 GTX 1050 Ti 4GB card with 32GB DDR4 on a single server Work Station. In the first part object detection \cite{28redmon2018yolov3} is used. The output for a frame consists of three lists each representation are 84-dimensional vectors. YOLO v3 makes predictions at 3 scales, that square measure exactly given by down-sampling the size of the input image by thirty-two, sixteen and eight severally. 

In the second module, if the prediction confidence score is generated if more than 0.5 then the object is extracted from that frame. Then output of these three feature vectors and object category is matched with other frames of a given video. With confidence threshold 0.5 and Non-Maximal suppression threshold 0.4. 

In the third module, accumulate the frame sequence and generate summarized video. During the testing phase, the proposed method obtain the feats consist of FrameId, Indices, Class, ClassId, Confidence of predicted shot and score for each video shot. 

\subsection{Evaluation Metrics} In QC-DPP \cite{24sharghi2017query}, the mapping between the predicted summary and the ground-truths is proposed with bipartite graph (Bipartite) using weight matching. Also gives a similarity function between two video shots by using intersection-over-union (IOU) on corresponding ideas to calculate the performance. The IOU is defined using edge weights therefore the predicted summary and ground-truth belongs to different sides of bipartite graph. Precision (Pre), recall (Rec), and F1-score (F1) area unit computed as follows.
\newline \newline
S1= Total no of Frames in Summary.\newline
S2= Total no of Frames in Ground\_Truth.\newline
Dist\.Matrix= distance (Summary, Ground\_Truth)\newline
Pre = Bipartite(Dist\. Matrix)/S1\newline
Rec = Bipartite(Dist\. Matrix)/S2\newline
F1 = 2*Pre*Rec / (Pre + Rec)

\begin{table*}
\caption{Comparison of proposed work with previous work}
\includegraphics[scale=.6]{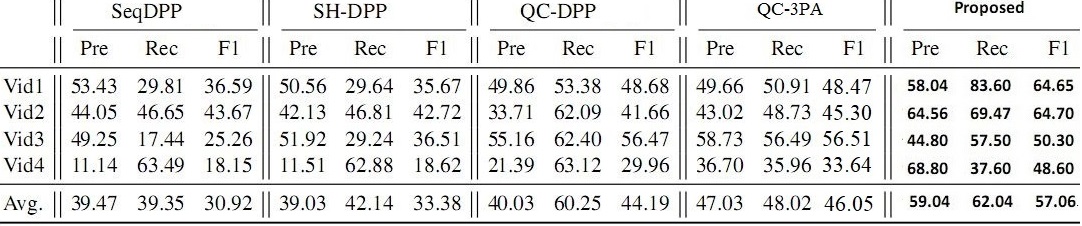}
\end{table*}
\subsection{ Quantitative Results}

The proposed approach is compared with all other frameworks which have been applied to UTE dataset. Precision, recall and F1-score Parameters are taken for comparison for all four videos are shown in Table 3 respectively. The proposed method achieved 57.06\% average F1-Score for UTE dataset. It can be observed that the proposed approach outperforms the existing approach by 11.01\%. Such substantial improvement in the performance indicates the superiority of the proposed method by using an Object detection method with other visual information and the image query. The rest four works are based on architecture which can take a long time to learn temporal relations among video shots and queries. However, the proposed work facilitates key short extraction using relation between video and image query.

To obtain a result comparison fast with ground truth values the proposed method has generated an auto-vector list along with every summarized video. So that  it  can compare this vector list with ground truth values. 

The results and analysis of query-conditioned video summarization proposed method using parameters (Pi) Precision, (Ri) Recall and (F1) F1-score. 

\begin{figure}
\includegraphics[scale=.4]{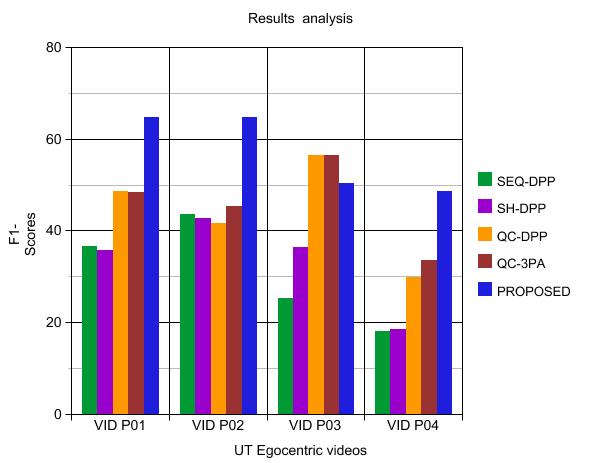}
\caption{Result analysis (F1-Score) of the proposed work with SEQ-DPP, SH-DPP, QC-DPP, QC-3PA}
\end{figure}

The process time is also calculated for all four videos and compared with actual timing of videos. Proposed method is 7.81 times less than actual time of video. Hence, by using this method we can summarize a video in very less time as compared to manually by watching.  

\begin{figure}
\includegraphics[scale=.78]{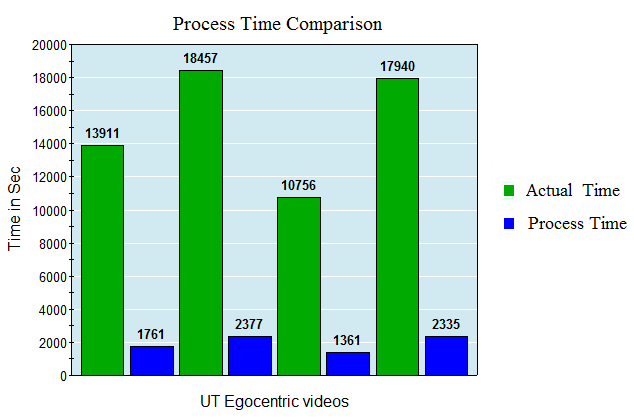}
\caption{ The average processing time on UTE dataset. (a) Actual time is the time length of videos. (b) Process time is the time taken by proposed method for generating video summary. }
\end{figure}

\subsection{Qualitative Results}

The visual results obtained by the proposed method are shown in Fig. 11. The proposed method uses an image query that contains two different objects Person and Car using ’OR’ operation. The x-axis represents the shot to video VIDP01. Ground-truth is denoted by blue lines for the given user query, while the green lines in the bottom represents predicted key shots of the proposed method. Note that predicted summaries can be related to one or more details given a user query. It can observe that compact and representative summaries can be find by the proposed method . 

\begin{figure}
\includegraphics[scale=.5]{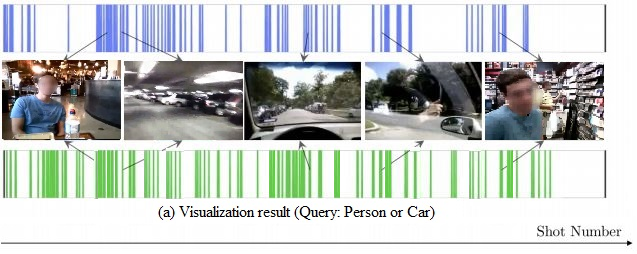}
\caption{ Visualization results for the proposed method with shot number on the x-axis in video VIDP01. Ground-truths, and  predicted key shots are shown in blue and green lines respectively. These results are for the query Person or Car.}
\end{figure}

\section{ Conclusion}
In the proposed work, a mathematical model is presented to minimize redundancy based on the similarity score between the query image and input video to obtain the summarized video. The mathematical model contains a method to calculate the summary variance to reduce redundancy between frames. A mathematical formula to calculate distance score between a query image and video frames is also presented. The presented work defined a loss function for obtaining a selection score for frames in a video. The proposed method assumes that the distribution of the keyframes is based on Gaussian distribution. Thus, an adaptive method to compute threshold is defined for the selection of keyframes using standard deviation. The Object-based Query Image (OQI) dataset is prepared for a selection of query images. The proposed model successfully resolved the issues of (i) user preference, (ii) recognize important frames and selecting that keyframe in daily life videos, with different illumination conditions. The proposed method achieved a 57.06\% average F1-Score for the UTE dataset. A video pre-processing process which makes use of very low-level features to efficiently locate undesirable frames then uses these to compute optimal segments. The processing time is 7.81 times less than the actual time of video Future work will be a focus on increasing local and global features to improve user subjectivity. 

\textbf{Conflicts of Interest:} The authors declare no conflict of interest.
\bibliographystyle{unsrt}       
\bibliography{bibfile} 

\end{document}